\documentclass[11pt]{article}

\usepackage[preprint]{acl}

\usepackage{times}
\usepackage{latexsym}

\usepackage[T1]{fontenc}

\usepackage[utf8]{inputenc}

\usepackage{microtype}

\usepackage{graphicx}
\usepackage{amsmath,amssymb}
\usepackage{booktabs}
\usepackage{multirow}
\usepackage{stfloats}
\usepackage{float}
\usepackage{xcolor}
\usepackage{tcolorbox}

\newcommand{\dice}{\textsc{DICE}}
\newcommand{\single}{\textsc{Single}}
\newcommand{\diceours}{\textsc{DICE}~(\textit{ours})}
\newcommand{\singlecite}{\textsc{Single}~(\citealp{zhang-etal-2025-diffusion})}
\newcommand{\latechunkdoc}{\textsc{LateChunk-Doc}}
\newcommand{\latechunkdoccite}{\textsc{LateChunk}~(\citealp{Gunther2024LateCC})}
\newcommand{\deltapos}[1]{\textcolor[HTML]{C62828}{#1}}
\newcommand{\deltaneg}[1]{#1}
\definecolor{caseheader}{HTML}{4F6FBF}
\definecolor{caseborder}{HTML}{6E86C9}
\definecolor{casebg}{HTML}{F4F7FC}
\newtcolorbox{casebox}[1]{
colback=casebg,
colframe=caseborder,
coltitle=white,
colbacktitle=caseheader,
arc=5pt,
boxrule=0.6pt,
left=6pt,
right=6pt,
top=6pt,
bottom=6pt,
title=#1,
fonttitle=\bfseries,
center title,
before upper={\small},
fontupper=\fontfamily{ptm}\selectfont,
}
\title{Lost in a Single Vector: Improving Long-Document Retrieval with Chunk Evidence Aggregation}

\author{
    Shanshan Lyu$^{1,2,3}$\thanks{~~Work performed primarily as a visiting student at the Institute of Computing Technology, Chinese Academy of Sciences; currently a student at Chongqing University.}
    \thanks{~~Authors from affiliation $^{2,3}$ are also affiliated with University of Chinese Academy of Sciences.} \quad
    Yiwei Wang$^{4}$ \quad
    Yujun Cai$^{5}$ \quad
    Jiafeng Guo$^{2,3}$\footnotemark[2]  \quad
    Shenghua Liu$^{2,3}$\footnotemark[2] \thanks{~~Corresponding author.}\quad\\
    $^1$Chongqing University \quad
    $^2$State Key Laboratory of AI Safety \\
    $^3$Institute of Computing Technology, Chinese Academy of Sciences \\
    $^4$University of California, Merced \quad
    $^5$University of Queensland \\
    {\small \texttt{shanshanlyu@stu.cqu.edu.cn}, \texttt{\{guojiafeng, liushenghua\}@ict.ac.cn}}
}

\makeatletter
\renewcommand\@makefntext[1]{%
  \noindent\makebox[-1pt][l]{\@makefnmark\hspace{0em}}#1}
\makeatother

\begin{document}
\maketitle

\begin{abstract}
Dense retrieval ranks one query vector against one document vector. On long documents, this interface can fail when a short but decisive span is weakened during document encoding before ranking. We study this failure mode as \emph{document-side early compression} and introduce the Evidence Dilution Index (EDI) to measure how far a document-level representation falls below the strongest chunk-level evidence within the same gold document. Guided by this view, we propose \dice{} (\textbf{D}ocument \textbf{I}nference via \textbf{C}hunk \textbf{E}vidence), a training-free document-side strategy that splits documents into chunks, encodes them independently with a frozen model, and aggregates them back into a single vector while preserving the standard one-query-one-document interface. On LongEmbed, \dice{} improves retrieval across four backbones, with the largest gains on slices beyond 4k tokens: for Dream, Passkey $>$4k rises from 30.0 to 90.0 and Needle $>$4k from 23.3 to 74.0. Across 12,779 filtered samples, \dice{} yields lower EDI than the single-vector baseline in 92.8\% of cases. These results establish document-level encoding as a practical and underexplored lever for long-document retrieval. Our code is available at \url{https://github.com/PunchlineAAAA/DICE}.
\end{abstract}

\section{Introduction}

Retrieval systems are increasingly asked to find answers inside long documents such as meeting transcripts, legal filings, and narrative texts, where the decisive evidence may occupy only a few sentences amid thousands of irrelevant tokens. Dense retrieval handles this by encoding each document into a single vector and ranking by query--document similarity. The interface is simple and scales to billions of documents, but it creates a representational bottleneck: one vector must summarize an entire document, even when relevance turns on a short local span.

\begin{figure}[t]
    \centering
    \includegraphics[width=\linewidth]{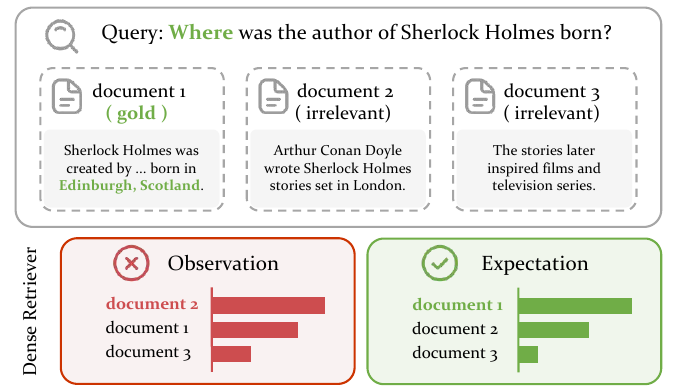}
    \caption{A motivating long-document retrieval case. The gold document contains a decisive local span, yet single-vector encoding under-ranks it because the evidence is diluted before retrieval.}
    \label{fig:case-study}
\end{figure}

This bottleneck is not merely a capacity problem. Modern encoders routinely support context windows of 4k--32k tokens, yet retrieval quality still degrades on long documents~\citep{zhu-etal-2024-longembed}, and models systematically under-utilize information away from salient positions~\citep{liu-etal-2024-lost}. The underlying cause is representational: when a query is resolved by a few decisive sentences embedded in a long document, encoding the whole document compresses the relevant evidence together with a large volume of irrelevant context before any query comparison is made. We refer to this failure mode as \textbf{document-side early compression}. As Figure~\ref{fig:case-study} illustrates, the gold document may contain the answer-bearing span and still receive a weak document-level score, because the evidence is diluted before retrieval ever happens.

\begin{figure*}[t]
    \centering
    \includegraphics[width=\linewidth]{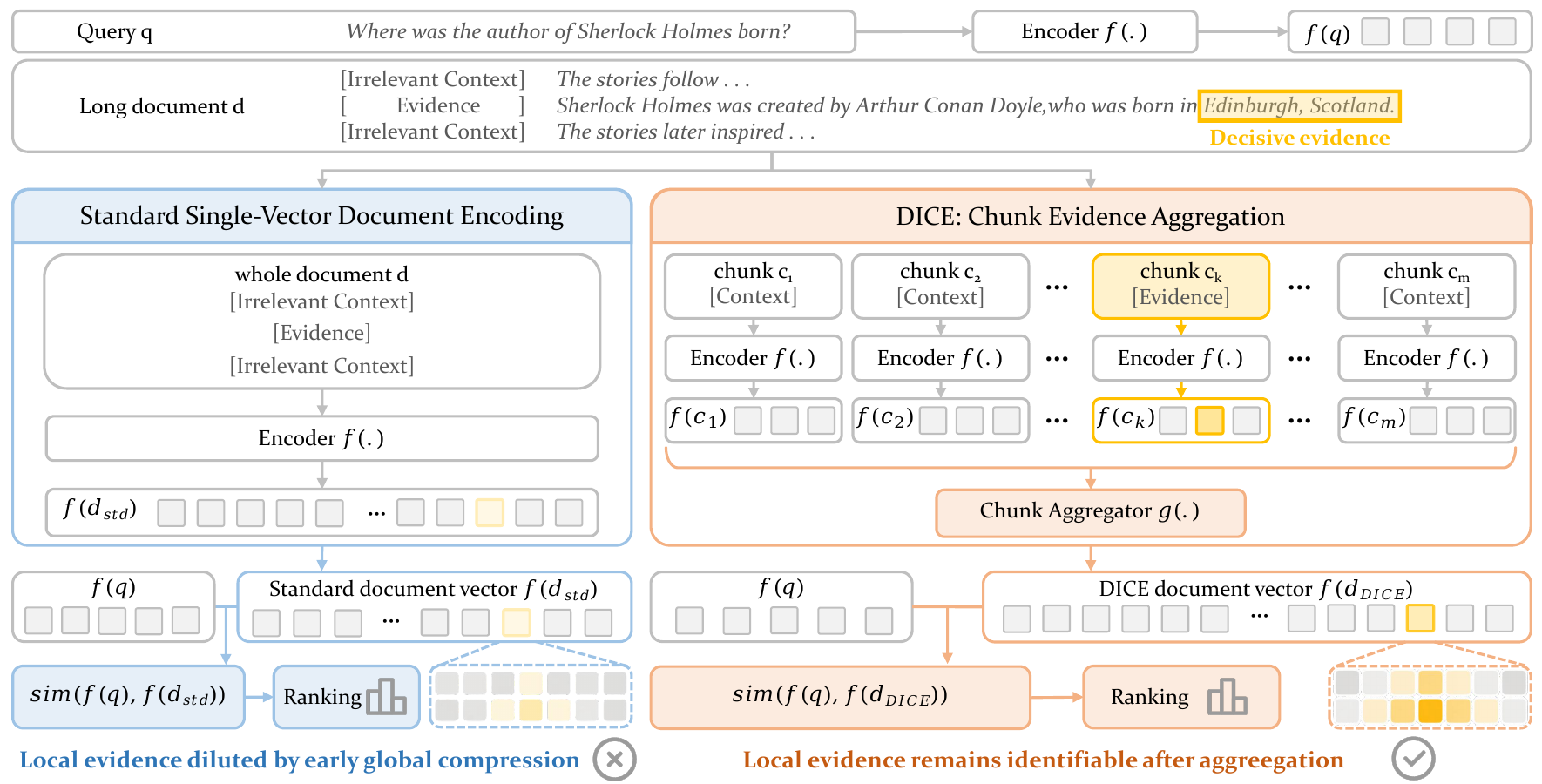}
    \caption{Overview of \dice{}. Only the document side changes: chunks are encoded independently with local positions and aggregated into one vector. The query path and the retrieval interface remain unchanged.}
    \label{fig:method}
\end{figure*}

This failure mode is difficult to address within the standard single-vector interface, and equally difficult to measure directly. Retrieval quality degrades with context length~\citep{zhu-etal-2024-longembed}, but that degradation conflates many factors and does not isolate the contribution of document-side compression. Methods that resolve the problem more directly, such as passage retrieval and late-interaction scoring~\citep{khattab2020colbert,santhanam-etal-2022-colbertv2}, do so by changing the retrieval unit or the scoring interface entirely. This leaves the document encoding side as a practical and underexplored degree of freedom.

We take two steps to exploit it. First, we introduce the \textbf{Evidence Dilution Index} (EDI), a per-sample diagnostic that compares a document-level similarity score against the strongest chunk-level evidence within the same gold document, turning early compression from an intuition into a measurable property. Second, guided by this diagnostic, we propose \dice{} (\textbf{D}ocument \textbf{I}nference via \textbf{C}hunk \textbf{E}vidence), a training-free strategy that splits a document into chunks, encodes each chunk independently with local position indices, and aggregates the resulting embeddings back into a single document vector (Figure~\ref{fig:method}). Query encoding is unchanged, so the retrieval interface remains standard one-query-one-document retrieval.

We evaluate \dice{} on LongEmbed~\citep{zhu-etal-2024-longembed} across four backbones spanning diffusion (Dream) and autoregressive (Mistral~\citep{Jiang2023Mistral7}, Llama3~\citep{Dubey2024TheL3}, Qwen~\citep{yang2024qwen25}) architectures. \dice{} consistently improves over single-vector baselines, with the largest gains on the hardest long-context slices: for Dream, Passkey $>$4k rises from 30.0 to 90.0 and Needle $>$4k from 23.3 to 74.0. Ablations identify chunk granularity as the key design factor, and EDI analysis explains the mechanism: across 12,779 filtered samples, \dice{} yields lower EDI than the single-vector baseline in 92.8\% of cases. FollowIR~\citep{Weller2024FollowIREA} provides secondary transfer evidence beyond explicitly long-document benchmarks, though the best chunk configuration is task dependent.

Our contributions are:
\begin{itemize}
    \item We identify document-side early compression as a failure mode in long-document dense retrieval and introduce EDI to quantify it.
    \item We propose \dice{}, a training-free document-side remedy that preserves the standard one-query-one-document retrieval interface.
    \item Across four backbones and two benchmarks, we show that \dice{} improves long-document retrieval and that these gains coincide with systematically lower evidence dilution.
\end{itemize}

\section{Evidence Dilution in Long-Document Retrieval}
\label{sec:edi-framework}

\subsection{Document-Side Early Compression}

Single-vector document encoding fails when relevance is localized. Given a query $q$ and a long document $d$, a dense retrieval encoder $f(\cdot)$ must compress the entire document into one vector before retrieval begins. Documents are then ranked by cosine similarity:
\begin{equation}
    s(q,d) = \operatorname{sim}(f(q), f(d)),
    \mathbf{d}_{\mathrm{single}} = f(d).
\end{equation}
When $d$ contains a short evidence span $e$ embedded in a large volume of non-relevant context $c$, the encoding $f(e,c)$ is pulled away from the representation of the decisive evidence before the similarity score is ever computed. The document may contain the answer and still rank poorly, because the document-level vector under-represents the short local span that actually resolves the query. Modern encoders routinely process 4k--32k tokens, so this is not primarily a failure of context-window capacity. Rather, it is a failure of representation: $\mathbf{d}_{\mathrm{single}}$ can lie far from the query in embedding space even though the gold document contains the answer-bearing evidence. Figure~\ref{fig:case-study} gives the motivating intuition. The analytical question is therefore not merely whether relevant evidence exists in the document, but whether single-vector compression preserves that evidence strongly enough to influence ranking.

\subsection{From Local Evidence to EDI}
\label{sec:edi-metric}

To make this failure mode measurable, we compare document-level representations against chunk-level evidence inside the same gold document. For a gold document split into $M$ chunks, let $a_j$ be the query--chunk similarity, $m = \max_j a_j$ the strongest local evidence, and $\bar{a}$ the mean chunk similarity. The max-over-chunks score is used here only as a local-evidence oracle for analysis; it is not intended as a deployment-equivalent baseline.

\smallskip
\noindent
\textbf{Evidence Concentration (EC).}
$\text{EC} = \dfrac{m - \bar{a}}{|m| + \epsilon}$.
EC captures whether query relevance is sharply localized in a few chunks (high EC) or spread more uniformly across the document (low EC). This tells us whether a sample is structurally susceptible to evidence dilution.

\noindent
\textbf{Evidence Dilution Index (EDI).}
$\text{EDI}(E) = \dfrac{m - s_E}{m - \bar{a} + \epsilon}$,
where $s_E = \text{sim}(q, \mathbf{d}_E)$ is the document-level similarity score under encoding method $E$. EDI measures how far the document vector falls below the strongest chunk-level evidence. Lower EDI indicates that the document representation remains closer to the most relevant local span; negative EDI means the document vector exceeds the single-chunk oracle by integrating signal across multiple chunks.

\smallskip
Both metrics use the evidence margin $m - \bar{a}$ as a normalizer. When this margin is near zero, the ratio can become unstable; this occurs most often in synthetic settings where nearly all chunks are irrelevant to the query. We therefore filter samples with $m - \bar{a} < 0.01$ and report median statistics throughout.

\subsection{Implications for Document Encoding}

This framework turns early compression from an intuition into a concrete design target. If a single document vector systematically falls below the strongest chunk-level evidence, then a natural remedy is to delay compression: preserve local evidence first, and aggregate only afterward. \dice{} follows exactly this strategy. It does not alter the query encoder or the retrieval interface; it changes only how the document vector is constructed so that decisive local evidence is less likely to be washed out before ranking.

\section{Method}

Guided by the evidence-dilution framework in Section~\ref{sec:edi-framework}, we now describe \dice{}.

\subsection{\dice{} Document Encoding}

\dice{} addresses the problem at the document encoding stage. The key idea is to delay compression: instead of encoding the whole document at once, we encode it in chunks and aggregate only afterward.

\paragraph{Chunking.} Given a document $d$, we tokenize it and split the token sequence into chunks of size $k$ with optional overlap $o$. The split is performed in token space, preserving token identities and boundaries. Unless otherwise noted, we use non-overlapping chunks ($o=0$) and vary $k$ to study chunk granularity.

\paragraph{Local position encoding.} Each chunk $d^{(j)}$ is encoded independently by the frozen encoder $f$. Position indices are reset to start from zero within each chunk, rather than being inherited from the original document:
\begin{equation}
    \mathbf{h}_j = f(\mathbf{x}^{(j)}, \mathbf{p}^{(j)}),
\end{equation}
where $\mathbf{x}^{(j)}$ are the token ids of chunk $d^{(j)}$ and $\mathbf{p}^{(j)}$ is the local position sequence that starts from zero within each chunk. Local positions ensure that each chunk is processed as a self-contained context window, unaffected by its offset in the source document. Query encoding is unchanged because our goal is to isolate document-side compression: queries are short relative to documents, so chunking them would add complexity on the query side while changing the retrieval interface we are trying to preserve.

\paragraph{Aggregation.} Chunk embeddings are fused into one document vector through a query-independent aggregation function $g$:
\begin{equation}
    \mathbf{d}_{\dice} = g(\mathbf{h}_1,\mathbf{h}_2,\ldots,\mathbf{h}_m).
\end{equation}
Aggregation is performed before the query is seen, so it cannot use query--chunk interactions. We therefore focus on a small family of simple, query-independent pooling rules that span the main design choices in this setting: preserving all chunks equally, emphasizing high-activation chunks, or selecting only a subset of chunks. Specifically, we study mean pooling, max pooling, and top-$k$ pooling by embedding norm. The default strategy is mean pooling, $g_{\mathrm{mean}} = \frac{1}{m}\sum_{j=1}^{m}\mathbf{h}_j$, which preserves evidence from every chunk equally. Max pooling, $g_{\mathrm{max}}$, tests a more selective alternative by taking the elementwise maximum across chunks. The norm-based variant uses $\|\mathbf{h}_j\|_2$ as a query-independent proxy for chunk salience: $g_{\mathrm{topk}}$ averages the top-$k$ highest-norm chunks. All variants are query-independent by construction and are evaluated as ablations in Section~\ref{sec:agg-ablation}.

\paragraph{Retrieval interface.} After aggregation, retrieval proceeds exactly as in the single-vector baseline: $s(q,d) = \operatorname{sim}(f(q), \mathbf{d}_{\dice})$. \dice{} does not retrieve chunks, increase the number of corpus entries, or add a second-stage scorer. The change is entirely on the document encoding side, and any difference in ranking comes only from the document representation.

\section{Experiments}

\subsection{Setup}

\paragraph{Tasks and Metrics.}
Our main benchmark is LongEmbed \citep{zhu-etal-2024-longembed}, which contains two synthetic tasks, Passkey and Needle, and four real retrieval tasks, NarrativeQA, QMSum, SummScreenFD, and WikiMQA. We report Hit@1 (\%) for Passkey and Needle, split into contexts up to 4k tokens and beyond 4k tokens. We report nDCG@10 for the real tasks. The split metrics are macro-averages over the corresponding LongEmbed length slices.

All evaluations use the MTEB framework~\citep{muennighoff-etal-2023-mteb}. We additionally evaluate Dream on FollowIR~\citep{Weller2024FollowIREA} (News21InstructionRetrieval, Core17InstructionRetrieval, Robust04Instruction-Retrieval), using nDCG@5 for News21 and MAP@1000 for Core17 and Robust04, with p-MRR as an auxiliary metric.

\paragraph{Comparison scope.}
Our main comparisons focus on \emph{deployment-equivalent} settings: one stored vector per document, unchanged query encoding, and the same first-stage query--document scoring interface. Under this scope, \single{} and the alternative chunk-aggregation rules in Section~\ref{sec:agg-ablation} are the direct baselines. We discuss passage retrieval, ColBERT-style late interaction, and late chunking as adjacent references rather than like-for-like competitors, since they change the retrieval unit, the scoring mechanism, or both.

\paragraph{Models.}
The primary backbone is a Dream-based diffusion embedder~\citep{zhang-etal-2025-diffusion}, built on the Dream 7B language model~\citep{Ye2025Dream7D}. To test whether the effect is specific to diffusion architectures, we also evaluate three autoregressive backbones converted to bidirectional embedders~\citep{BehnamGhader2024LLM2VecLL}: Llama3 8B~\citep{Dubey2024TheL3}, Mistral 7B~\citep{Jiang2023Mistral7}, and Qwen2.5 7B~\citep{yang2024qwen25}. In all comparisons, the encoder weights are frozen between the \single{} and \dice{} settings.

\paragraph{Implementation.}
\dice{} is implemented as document-side logic within a unified evaluation wrapper, built on an extended version of the LLM2Vec codebase~\citep{BehnamGhader2024LLM2VecLL} that supports both diffusion and autoregressive backbones. Query inputs are routed through ordinary encoding; document inputs are routed through either single-vector encoding or chunk aggregation depending on the configuration. In the chunked path, document windows are formed directly from token ids and each window is encoded with local positional indices. Unless otherwise stated, \dice{} uses chunk size 1024, no overlap, mean aggregation, bfloat16 inference, and cosine similarity. All evaluation runs use a single NVIDIA A6000 GPU with matched \single{} and \dice{} settings.

\subsection{Main Results}

Table~\ref{tab:longembed-main} reports the core result: across four backbones, replacing single-vector document encoding with \dice{} improves LongEmbed, with the largest gains concentrated on the hard long-context synthetic slices, exactly where the early-compression hypothesis predicts. For Dream, Passkey $>$4k increases by 60.0 points and Needle $>$4k by 50.7 points. The same trend holds across architectures: Mistral and Qwen show similarly large Passkey $>$4k gains, while Llama3 benefits most on NarrativeQA, SummScreenFD, and WikiMQA. The few regressions are small and occur where documents are shorter or relevance is less localized.

As a Dream-side adjacent reference, we additionally test an adapted \latechunkdoc{} variant based on Late Chunking~\citep{Gunther2024LateCC}. On Dream, \latechunkdoc{} reaches 81.51 average points, substantially above direct single-vector encoding (63.41) and close to \dice{} (81.92). We therefore treat this result as a supplementary design reference rather than a primary baseline (\hyperref[app:latechunk-doc]{Appendix C.1}).

\begin{table*}[t]
  \centering
  \small
  \caption{Main LongEmbed results. \dice{} uses chunk size 1024 with mean aggregation. Avg.\ is the unweighted mean over eight metrics, including the two omitted synthetic $\leq$4k columns.}
  \label{tab:longembed-main}
  \begin{tabular*}{\textwidth}{@{\extracolsep{\fill}}llcccccccc}
    \toprule
    Backbone & Method & Avg. & \multicolumn{2}{c}{Synthetic (Hit@1, \%)} & \multicolumn{4}{c}{Real (nDCG@10)} \\
    \cmidrule(lr){4-5} \cmidrule(lr){6-9}
     &  &  & Pk$>$4k & Nd$>$4k & NQA & QMS & SFD & WQA \\
    \midrule
    Llama3 & \singlecite{} & 34.64 & 3.33 & 7.33 & 18.06 & \textbf{29.24} & 77.45 & 40.08 \\
     & \diceours{} & \textbf{50.56} & \textbf{35.33} & \textbf{25.33} & \textbf{45.64} & 27.59 & \textbf{93.00} & \textbf{61.16} \\
     & $\Delta$ & \deltapos{+15.92} & \deltapos{+32.00} & \deltapos{+18.00} & \deltapos{+27.58} & \deltaneg{-1.64} & \deltapos{+15.55} & \deltapos{+21.08} \\
    \midrule
    Qwen & \singlecite{} & 43.80 & 8.67 & 9.33 & 19.51 & 35.82 & 87.89 & 51.95 \\
     & \diceours{} & \textbf{59.06} & \textbf{66.67} & \textbf{20.00} & \textbf{35.25} & \textbf{42.06} & \textbf{91.83} & \textbf{61.51} \\
     & $\Delta$ & \deltapos{+15.27} & \deltapos{+58.00} & \deltapos{+10.67} & \deltapos{+15.75} & \deltapos{+6.24} & \deltapos{+3.94} & \deltapos{+9.56} \\
    \midrule
    Mistral & \singlecite{} & 57.24 & 30.00 & 18.00 & 33.10 & 43.50 & 92.64 & 61.08 \\
     & \diceours{} & \textbf{74.38} & \textbf{86.00} & \textbf{58.67} & \textbf{55.73} & \textbf{50.79} & \textbf{96.48} & \textbf{71.38} \\
     & $\Delta$ & \deltapos{+17.14} & \deltapos{+56.00} & \deltapos{+40.67} & \deltapos{+22.63} & \deltapos{+7.29} & \deltapos{+3.84} & \deltapos{+10.31} \\
    \midrule
    Dream & \singlecite{} & 63.41 & 30.00 & 23.33 & 43.63 & 43.93 & 97.77 & 79.43 \\
     & \diceours{} & \textbf{81.92} & \textbf{90.00} & \textbf{74.00} & \textbf{65.01} & \textbf{50.54} & \textbf{98.53} & \textbf{86.04} \\
     & $\Delta$ & \deltapos{+18.50} & \deltapos{+60.00} & \deltapos{+50.67} & \deltapos{+21.39} & \deltapos{+6.61} & \deltapos{+0.76} & \deltapos{+6.61} \\
    \bottomrule
  \end{tabular*}
\end{table*}

\begin{table*}[t]
  \centering
  \caption{Dream ablation on LongEmbed. The top block varies chunk size with mean aggregation, and the bottom block fixes chunk size to 1024 while varying the aggregation rule. ``---'' denotes the \single{} baseline, and bold indicates the best per-column value.}
  \label{tab:ablation}
  \small
  \begin{tabular*}{\textwidth}{@{\extracolsep{\fill}}cclcccccccc}
    \toprule
    Chunk & Agg. & Avg & \multicolumn{4}{c}{Synthetic (Hit@1, \%)} & \multicolumn{4}{c}{Real (nDCG@10)} \\
    \cmidrule(lr){4-7} \cmidrule(lr){8-11}
     &  &  & Pk$\leq$4k & Pk$>$4k & Nd$\leq$4k & Nd$>$4k & NQA & QMS & SFD & WQA \\
    \midrule
    --- & --- & 63.41 & \textbf{100.0} & 30.00 & 89.20 & 23.33 & 43.63 & 43.93 & 97.77 & 79.43 \\
    128 & mean & 61.21 & 48.80 & 33.33 & 71.20 & 56.00 & 60.41 & 44.98 & 97.74 & 77.22 \\
    256 & mean & 69.83 & 75.20 & 34.00 & 84.80 & \textbf{74.67} & 63.09 & 47.43 & 98.36 & 81.12 \\
    512 & mean & 76.30 & 88.80 & 66.67 & 84.40 & 74.00 & 64.17 & 49.60 & \textbf{98.66} & 84.07 \\
    1024 & mean & \textbf{81.92} & 98.80 & 90.00 & \textbf{92.40} & 74.00 & \textbf{65.01} & \textbf{50.54} & 98.53 & \textbf{86.04} \\
    \midrule
    1024 & max & 64.63 & 99.20 & \textbf{100.0} & 89.20 & 44.67 & 26.65 & 25.39 & 88.40 & 43.53 \\
    1024 & topk & 56.65 & 86.80 & 9.33 & 88.00 & 18.00 & 49.60 & 39.99 & 95.39 & 66.09 \\
    \bottomrule
  \end{tabular*}
\end{table*}

\subsection{Analysis}
\label{sec:analysis}

\begin{figure}[t]
    \centering
    \includegraphics[width=\linewidth]{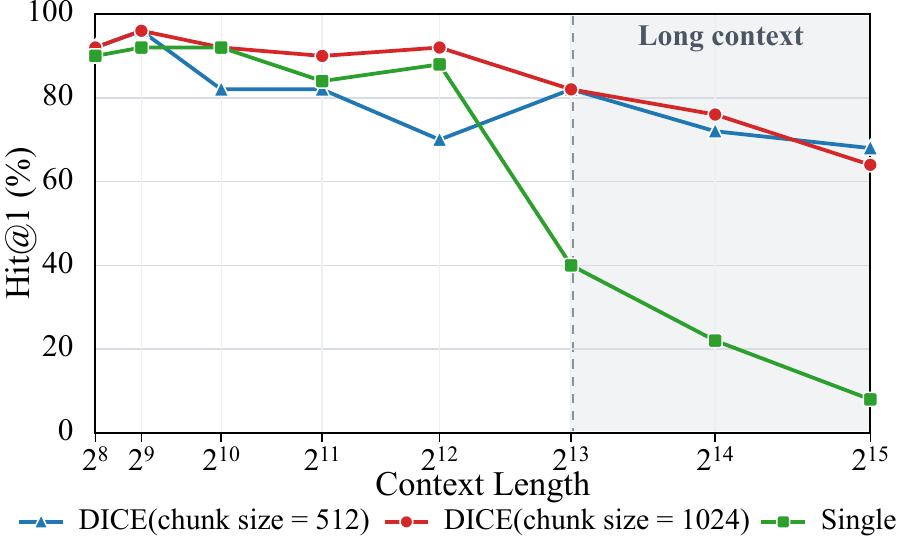}
    \caption{Needle Hit@1 (\%) by context length on Dream. Chunk-level encoding prevents the sharp long-context degradation observed under single-vector document compression.}
    \label{fig:pk-nd}
\end{figure}

\paragraph{Length-Stratified Needle.}
Figure~\ref{fig:pk-nd} breaks Needle Hit@1 (\%) down by context length for Dream.
At 4k tokens, the single-vector encoder matches the chunked variants; by 32k, it collapses to 0.08 while chunk-512 and chunk-1024 both remain above 0.64.
The gap widens monotonically beyond 4k, indicating that early compression, rather than context-window capacity alone, is a major bottleneck in long-document retrieval.

\paragraph{Ablation Analysis.}
\label{sec:agg-ablation}

Table~\ref{tab:ablation} reports a full ablation of the two design dimensions in \dice{}, chunk granularity and aggregation strategy, on the Dream backbone. All runs use non-overlapping chunks, bfloat16 inference, and cosine similarity; only the ablated variable changes.

\begin{figure}[t]
    \centering
    \includegraphics[width=\linewidth]{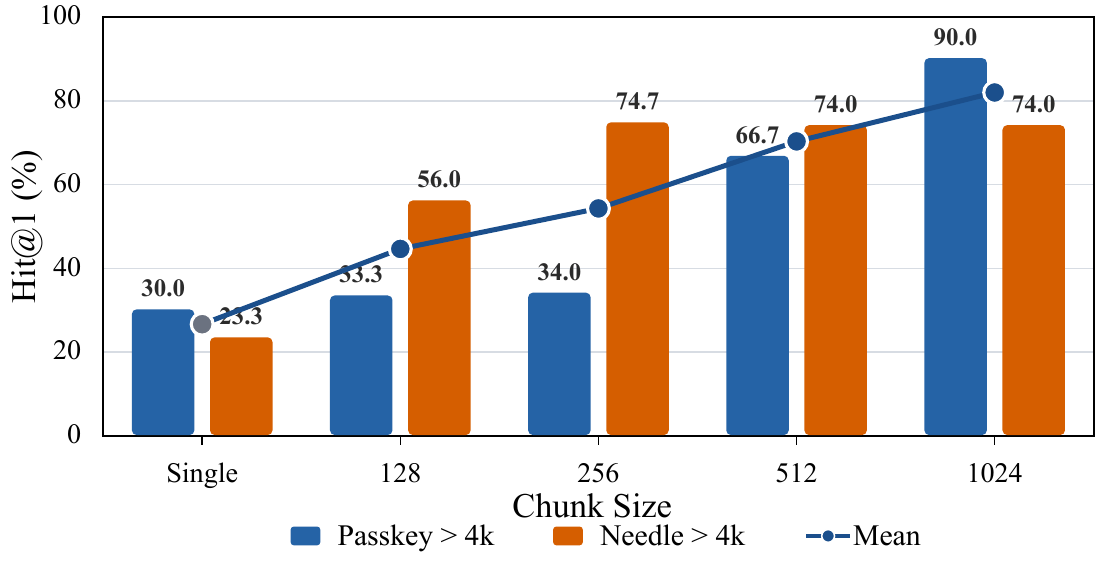}
    \caption{Dream chunk-size ablation on LongEmbed. Bars show Passkey $>$4k and Needle $>$4k Hit@1 (\%), and the line shows their mean. Larger chunk sizes consistently improve the hardest long-context slices, with chunk size 1024 giving the highest hard-slice mean.}
    \label{fig:chunk-size}
\end{figure}

Two clear patterns emerge. First, \textbf{chunk granularity matters decisively}. At 128 tokens, \dice{} underperforms the \single{} baseline on average (61.21 vs.\ 63.41), suggesting that overly small chunks fail to preserve enough useful local context. From 256 upward, the overall average rises monotonically, while the hard long-context slices (Passkey $>$4k and Needle $>$4k) improve even more sharply. Figure~\ref{fig:chunk-size} visualizes these two hardest slices together with their mean, while Figure~\ref{fig:pk-nd} shows the corresponding length-stratified Needle breakdown. Chunk size 1024 achieves the best average (81.92) and the best or near-best score on seven of eight metrics. The same ordering largely persists under Hit@$n$ for $n \in \{2,3,5,10\}$ on the synthetic tasks (\hyperref[app:topk-synth]{Appendix B.2}) and on Mistral (avg.\ rises from 52.77 at chunk-128 to 74.38 at chunk-1024; \hyperref[app:mistral-chunk]{Appendix B.1}).

Second, \textbf{more sophisticated aggregation is not better}. Max pooling reaches 100.0 on Passkey $>$4k, the highest of any setting, but collapses on real tasks: NQA drops from 65.01 to 26.65, QMS from 50.54 to 25.39, and WQA from 86.04 to 43.53. The norm-based variant (topk) also underperforms mean pooling across all metrics and often falls below even the \single{} baseline. The gains therefore come from preserving enough local context and aggregating conservatively, not from sophisticated chunk selection. We use mean pooling and chunk size 1024 as the default throughout the main results.

We also probed an alternative position scheme that preserves each chunk's absolute document offset instead of resetting positions locally. Across the completed backbones, this variant never improved over local reset: Avg.\ drops from 81.92 to 81.85 on Dream, from 50.56 to 43.19 on Llama3, and from 74.38 to 73.74 on Mistral (\hyperref[app:abspos]{Appendix C.2}). We therefore keep local reset as the default. We do not report Qwen for this probe because absolute offsets pushed chunk-level position ids beyond the backbone's supported range on very long inputs, exposing a practical robustness weakness that local reset avoids.

\begin{figure}[t]
    \centering
    \includegraphics[width=\linewidth]{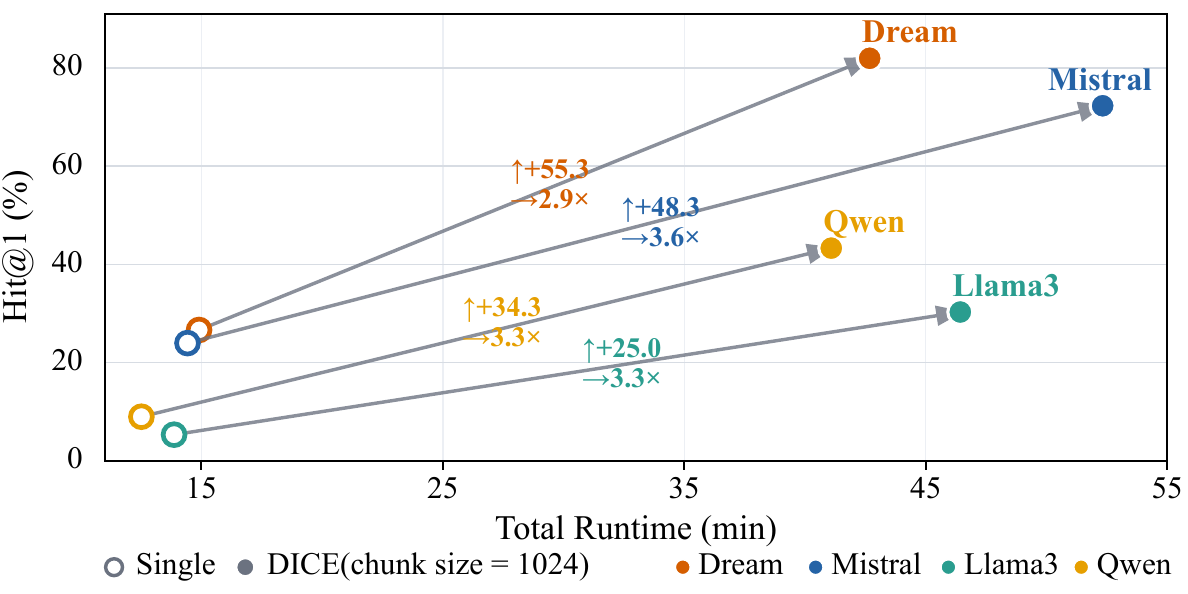}
    \caption{Quality-cost trade-off on LongEmbed. Quality is the mean of Passkey $>$4k and Needle $>$4k Hit@1 (\%), and arrows point from \single{} to \dice{} for each backbone.}
    \label{fig:tradeoff}
\end{figure}

\subsection{Deployment Trade-offs}

The analyses above establish the main mechanism claim. We now turn to two secondary deployment considerations that do not change that interpretation but matter for practical use: overlap and document-side encoding cost.

\paragraph{Effect of Overlap.}
Introducing token overlap between adjacent chunks (chunk size 1024) does not improve average performance over the non-overlapping baseline. With 256-token overlap, the average score drops from 81.92 to 70.79, largely because Passkey $>$4k falls from 90.0 to 46.7. Increasing overlap to 512 tokens partially recovers the average (80.78) and improves Passkey $>$4k to 94.0 and Needle $>$4k to 79.3, but at a steep runtime cost (80~minutes for the full LongEmbed suite vs.\ 43 minutes without overlap). Since overlap increases document encoding time without a consistent average gain, we keep $o=0$ in the main setting.

\paragraph{Quality-Cost Trade-off.}
\dice{} is not an efficiency method: encoding multiple chunks per document increases document-side computation.

Figure~\ref{fig:tradeoff} plots the accuracy-runtime trade-off across backbones. The arrows show that \dice{} consistently moves models toward higher long-context Hit@1 (\%) at roughly three to four times the document-side runtime. The practical operating point depends on whether long-document recall justifies the added encoding cost, a trade-off that may be acceptable when documents are indexed offline and queried many times.

\subsection{EDI-Based Validation}

The ablations in Section~\ref{sec:analysis} establish that chunk granularity and conservative aggregation drive \dice{}'s gains, but they do not directly test our central hypothesis: that single-vector encoding systematically dilutes localized evidence. Using the EDI framework from Section~\ref{sec:edi-metric}, we now test that hypothesis directly. All analyses in this subsection use the Dream backbone with chunk size 1024 and mean aggregation. Here, the \emph{robust subset} means all LongEmbed samples that survive the margin filter from Section~\ref{sec:edi-metric}, i.e., $m - \bar{a} \geq 0.01$; for Dream, this leaves 12,779 samples in total. We report the overall comparison on this full subset, and then break out the four real tasks separately.

\begin{table}[t]
  \centering
  \caption{EDI comparison on the robust subset of LongEmbed on Dream with chunk size 1024 and mean aggregation. The robust subset consists of all samples with $m - \bar{a} \geq 0.01$, and DBR denotes the fraction of samples where $\text{EDI}(\dice{}) < \text{EDI}(\single{})$.}
  \label{tab:edi}
  \small
  \setlength{\tabcolsep}{4pt}
  \begin{tabular}{lccc}
    \toprule
    Split & \single{} EDI $\downarrow$ & \dice{} EDI $\downarrow$ & DBR (\%) $\uparrow$ \\
    \midrule
    Overall & \phantom{-}0.055 & $-0.709$ & 92.8\% \\
    NQA & \phantom{-}0.046 & $-0.737$ & 95.7\% \\
    QMS & $-0.131$ & $-1.020$ & 94.7\% \\
    SFD & $-0.009$ & $-0.693$ & 92.7\% \\
    WQA & \phantom{-}0.286 & \phantom{-}0.205 & 61.0\% \\
    \bottomrule
  \end{tabular}
\end{table}

Table~\ref{tab:edi} reports both the overall comparison and the real-task breakdown. Overall, \dice{} reduces median EDI from $0.055$ to $-0.709$ and yields lower EDI than \single{} in 92.8\% of samples. The same pattern holds strongly on NarrativeQA, QMSum, and SummScreenFD, where median EDI drops substantially and DBR remains above 92\%. WikiMQA shows a smaller but still positive reduction (0.286 to 0.205), indicating that the effect is not perfectly uniform across tasks but remains directionally consistent. A lightweight Mistral check in \hyperref[app:mistral-edi]{Appendix A.2} shows the same qualitative pattern, suggesting that EDI is not merely a Dream-specific diagnostic.

\begin{figure}[t]
    \centering
    \includegraphics[width=\linewidth]{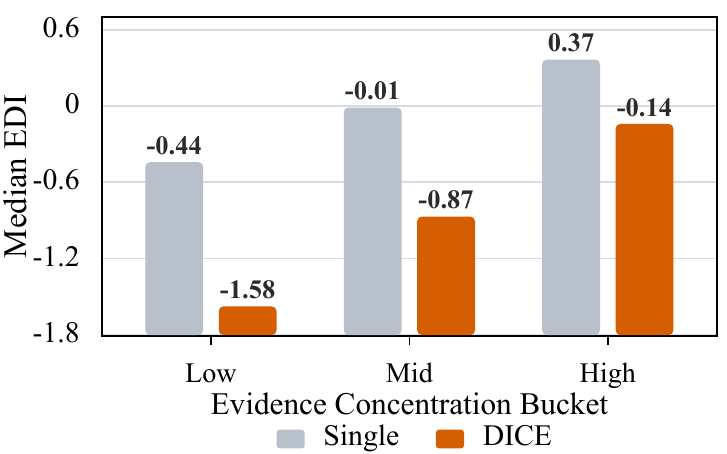}
    \caption{Median EDI across EC terciles on Dream for the LongEmbed robust subset. EDI rises with EC for both methods, but \dice{} remains consistently lower across all three buckets. The relative gap between \single{} and \dice{} is largest in the low-EC bucket.}
    \label{fig:edi-ec-bucket}
\end{figure}

Figure~\ref{fig:edi-ec-bucket} further decomposes the result by EC tercile (low $\leq 0.085$, mid $0.085$--$0.126$, high $> 0.126$; about 4,260 samples each). EDI rises with EC for both methods, indicating that sharply localized evidence is intrinsically harder to preserve in a single vector. At the same time, \dice{} remains better in every bucket (low: $-1.61$ vs.\ $-0.48$; mid: $-0.91$ vs.\ $-0.04$; high: $-0.16$ vs.\ $0.35$).

This breakdown sharpens the mechanism interpretation in two ways. First, it matches the length-stratified degradation observed earlier on the hardest long-context slices: as relevance becomes both more localized and more deeply buried, single-vector compression fails more often. Second, the relative gap is largest in the low-EC bucket, indicating that \dice{} helps not only when one chunk dominates, but also when useful evidence is distributed across multiple chunks.

\begin{table}[t]
  \centering
  \caption{FollowIR results on Dream. Improvements persist beyond LongEmbed, but the best chunk configuration remains task dependent.}
  \label{tab:followir}
  \small
  \setlength{\tabcolsep}{3.3pt}
  \begin{tabular}{lccc}
    \toprule
    Method & News & Core & Robust \\
     & nDCG@5 & MAP@1000 & MAP@1000 \\
    \midrule
    \single{} & 36.66 & 24.87 & 25.81 \\
    \dice{}-256 & 39.03 & 29.02 & 29.18 \\
    \dice{}-256+ov32 & \textbf{42.25} & 29.61 & 29.79 \\
    \dice{}-256+ov128 & 38.80 & \textbf{31.14} & \textbf{30.27} \\
    \dice{}-512 & 35.42 & 25.74 & 27.93 \\
    \bottomrule
  \end{tabular}
\end{table}

\subsection{Transfer to FollowIR}

As a secondary transfer check, Table~\ref{tab:followir} evaluates Dream on FollowIR. Document-side chunk aggregation consistently improves over the single-vector baseline, reaching 42.25 nDCG@5 on News21 (vs.\ 36.66), 31.14 MAP@1000 on Core17 (vs.\ 24.87), and 30.27 MAP@1000 on Robust04 (vs.\ 25.81). Unlike LongEmbed, however, the best FollowIR setting is not a single no-overlap chunk size: News21 benefits most from 256-token chunks with 32-token overlap, while Core17 and Robust04 prefer a larger 128-token overlap at the same chunk size. We therefore treat FollowIR as transfer evidence that the gains extend beyond explicitly long-document benchmarks, while highlighting that the best granularity remains benchmark dependent.

Taken together, the most reliable gains from \dice{} occur when documents are long and relevance is localized, and they come from preserving enough local context before document-level compression rather than from sophisticated chunk selection.

\section{Related Work}

\textbf{Dense retrieval and the single-vector bottleneck.}
Dense retrieval has steadily improved through stronger encoders while preserving a stable interface: one vector per retrieval unit. This trajectory runs from Sentence-BERT~\citep{reimers-gurevych-2019-sentence} and DPR~\citep{karpukhin-etal-2020-dense}, through contrastive embedders such as E5~\citep{Wang2022TextEB}, Contriever~\citep{Izacard2021UnsupervisedDI}, BGE~\citep{Xiao2023CPackPR}, and GTE~\citep{Li2023TowardsGT}, to LLM-based encoders such as LLM2Vec~\citep{BehnamGhader2024LLM2VecLL}, NV-Embed~\citep{ICLR2025_c4bf7338}, M3-Embedding~\citep{Chen2024M3EmbeddingMM}, GritLM~\citep{muennighoff2025gritlm}, and recent diffusion-based embedders~\citep{zhang-etal-2025-diffusion,Eslami2026DiffusionPretrainedDA}. The downside is that the same interface creates a bottleneck for long documents. Long-context studies show that even models with extended context windows under-use information distributed across long sequences~\citep{liu-etal-2024-lost,Hsieh2024RULERWT,zhang2024loogle}, and LongEmbed~\citep{zhu-etal-2024-longembed} makes this failure visible through length-stratified retrieval evaluation. \dice{} departs from this encoder-centric line of work by leaving the encoder frozen and changing only how the document vector is computed at inference time.

\textbf{Multi-vector and late-interaction methods.}
A parallel line of work relaxes the single-vector constraint. Passage retrieval, central to RAG pipelines~\citep{NEURIPS2020_6b493230}, splits documents into chunks before indexing and retrieves them independently, trading the clean one-vector interface for downstream synthesis complexity. Multi-vector late-interaction methods (ColBERT~\citep{khattab2020colbert}, ColBERTv2~\citep{santhanam-etal-2022-colbertv2}) retain token-level representations and compute fine-grained query--document alignment, improving fidelity at the cost of a more complex indexing and scoring pipeline. These methods demonstrate that preserving per-token or per-chunk evidence improves retrieval on long documents, but they require changes to both the indexing infrastructure and the retrieval interface. \dice{} borrows the intuition that fine-grained representations help, but preserves the standard one-query-one-document interface by aggregating chunk evidence back into a single document vector before retrieval.

\textbf{Chunk-level and segment-level representations.}
The closest line of work operates at the chunk or segment level. Late chunking~\citep{Gunther2024LateCC} contextualizes token embeddings through a long-context forward pass before pooling them into chunk representations for chunk-level retrieval. \dice{} shares the premise that evidence should be captured before global compression, but differs in both retrieval granularity and encoding path: it returns a single document vector and encodes chunks independently with local positions. SeDR~\citep{Chen2022SeDRSR} similarly models long documents at the segment level, but requires training a dedicated encoder, whereas \dice{} is training-free. More recently, \citet{Bhat2025RethinkingCS} show that chunk granularity substantially affects passage retrieval quality. Our ablations extend this observation to document-level representations, where the benefit of larger chunks persists when aggregation is kept query-independent and conservative. Learned compression of multi-vector representations into a single vector remains a promising next step, but it would move beyond the training-free setting studied here.

\section{Conclusion}

We framed long-document retrieval through the lens of document-side early compression and introduced EDI to measure it directly. Within that framework, \dice{} provides a training-free document-side remedy that preserves the standard single-vector retrieval interface while markedly improving retrieval when localized evidence would otherwise be diluted by document-level compression. Our ablations show that the central design choice is to preserve enough local context before aggregation, and that simple mean pooling is the most robust query-independent rule in this setting. More broadly, the results identify document-side encoding as a practical and underexplored degree of freedom for improving long-document retrieval.

\section*{Limitations}

\dice{} increases document-side encoding cost by roughly 3--4$\times$, a trade-off most acceptable when documents are indexed offline. The method aggregates chunk representations without seeing the query, which preserves the single-vector interface but foregoes the chance to dynamically emphasize query-relevant chunks. Finally, \dice{} still compresses each document into one vector, so it necessarily loses finer-grained evidence compared with multi-vector or passage-level retrieval. These three constraints (encoding cost, query-independent aggregation, and the single-vector ceiling) define the operating point on the accuracy--efficiency frontier that \dice{} occupies.

\section*{Ethical Statement}

This work does not involve human subjects, personally identifiable information, or private user data. All experiments are conducted on publicly available benchmark datasets and publicly released models, and we follow their licenses and terms of use. Our method is a training-free change to document encoding for retrieval and is intended to improve the reliability of long-document retrieval systems. However, better retrieval does not guarantee correct downstream decisions, and the best chunk configuration may not transfer uniformly across domains. We therefore recommend task-specific validation before deployment.

\bibliography{references}

\section*{Appendix A. Supplementary Mechanism Evidence}

This appendix strengthens the mechanism claim behind \dice{}. Whereas the main paper establishes the overall pattern quantitatively, the materials here answer two narrower questions: what evidence dilution looks like in a concrete retrieval example, and whether the EDI trend survives beyond the Dream backbone. The appendix is therefore meant to make the argument more interpretable.

\subsection*{A.1 QMSum Case}

Figure~\ref{fig:qmsum-case-appendix} gives a concrete example of the failure mode discussed in Section~\ref{sec:edi-framework}. The point is not that QMSum is special, but that the decisive evidence is visibly localized: a small set of short phrases resolves the query, yet the full-document \single{} representation ranks the gold document at 128. By contrast, \dice{}(chunk size = 1024) recovers rank 1, consistent with the claim that delaying document-side compression helps preserve concentrated evidence.

\begin{figure}[H]
    \centering
    \begin{casebox}{Localized Evidence in QMSum}
    \textbf{Query.} ``Industrial Designer thought the meeting was not friendly to the brainstorming. The restriction was not about the atmosphere but related to the actual environment and the limited time for discussion. Besides, the interaction was structured, meaning each individual took charge of one particular task without enough collaboration between each other. Also, communication through email was inefficient.''\par\medskip
    \textbf{Localized evidence in the gold document.} ``the meetings ... are more brainstorming sessions than meetings''; ``the room not being ... very friendly''; ``the time given also restricts''; ``the interactions are very structured''; ``each individual is structured to one particular task''; ``it just comes back to us so slow in the email''; ``they don't support collaboration.''\par\medskip
    \textbf{Gold document rank.} \single{}: 128 \hspace{1em} \dice{}: 1
    \end{casebox}
    \caption{A QMSum example with highly localized evidence. A short discussion segment contains the decisive evidence needed to recover the gold document.}
    \label{fig:qmsum-case-appendix}
\end{figure}

\subsection*{A.2 Cross-Backbone EDI Check}
\label{app:mistral-edi}

To complement the Dream-based main analysis, we report a lightweight Mistral check under the same robust-subset criterion. The purpose is not to replicate every analysis from Section~\ref{sec:analysis}, but to test whether the same mechanism appears under a different encoder family. Table~\ref{tab:mistral-chunk-size} first verifies that the chunk-size trend itself transfers to Mistral, and Table~\ref{tab:mistral-edi} then asks whether that transfer is accompanied by the same direction of EDI improvement.

\begin{table*}[t]
  \centering
  \caption{Mistral chunk-size sweep on LongEmbed. Synthetic scores are Hit@1 (\%), and real-task scores are nDCG@10. The same chunk-granularity trend observed on Dream remains visible on Mistral.}
  \label{tab:mistral-chunk-size}
  \small
  \begin{tabular*}{\textwidth}{@{\extracolsep{\fill}}lccccccccc}
    \toprule
    Chunk & Avg & Pk$\leq$4k & Pk$>$4k & Nd$\leq$4k & Nd$>$4k & NQA & QMS & SFD & WQA \\
    \midrule
    --- (\single{}) & 57.24 & 99.20 & 30.00 & 80.40 & 18.00 & 33.10 & 43.50 & 92.64 & 61.08 \\
    128          & 52.77 & 37.20 & 18.00 & 75.60 & 42.00 & 51.58 & 42.35 & 91.97 & 63.48 \\
    256          & 58.66 & 51.60 & 29.33 & 76.00 & 52.67 & 53.35 & 46.06 & 94.24 & 66.05 \\
    512          & 65.65 & 80.00 & 44.67 & 79.20 & 54.67 & 54.52 & 48.80 & 95.43 & 67.89 \\
    1024         & 74.38 & 96.00 & 86.00 & 80.00 & 58.67 & 55.73 & 50.79 & 96.48 & 71.38 \\
    \bottomrule
  \end{tabular*}
\end{table*}

\begin{table}[t]
  \centering
  \caption{Mistral EDI on the robust subset. Lower EDI and high DBR replicate the main qualitative pattern beyond Dream.}
  \label{tab:mistral-edi}
  \small
  \setlength{\tabcolsep}{10pt}
  \begin{tabular}{lccc}
    \toprule
    Split & \single{} $\downarrow$ & \dice{} $\downarrow$ & DBR (\%) $\uparrow$ \\
    \midrule
    Overall & \phantom{-}0.404 & $-0.361$ & 91.8\% \\
    NQA & \phantom{-}0.415 & $-0.407$ & 94.6\% \\
    QMS & \phantom{-}0.311 & $-0.394$ & 93.1\% \\
    SFD & \phantom{-}0.478 & $-0.228$ & 92.2\% \\
    WQA & \phantom{-}0.489 & \phantom{-}0.076 & 87.7\% \\
    \bottomrule
  \end{tabular}
\end{table}

Taken together, the two Mistral tables support the same interpretation as the Dream results. Larger chunks again perform better than overly fine chunking, and the best-performing configuration also yields markedly lower EDI and high DBR on every real-task split. The magnitudes differ from Dream, especially on WikiMQA, but the qualitative alignment between better retrieval and lower dilution remains intact.

\section*{Appendix B. Supplementary Retrieval Analyses}

This appendix extends the retrieval-side analysis behind the main ablation story. Its goal is not to introduce new claims, but to test whether the two most important empirical patterns from Section~\ref{sec:analysis} are robust under additional views: the chunk-size trend across backbones, and the long-context ranking improvements beyond Hit@1.

\begin{table*}[t]
  \centering
  \caption{Synthetic Hit@$k$ (\%) on Dream for the $>$4k slices. The ranking gains from \dice{} persist beyond Hit@1 across broader top-$k$ cutoffs.}
  \label{tab:topk-synth}
  \small
  \setlength{\tabcolsep}{4pt}
  \begin{tabular*}{\textwidth}{@{\extracolsep{\fill}}lcccccccccc}
    \toprule
    \multirow{2}{*}{Method} & \multicolumn{5}{c}{Passkey $>$4k} & \multicolumn{5}{c}{Needle $>$4k} \\
    \cmidrule(lr){2-6} \cmidrule(lr){7-11}
     & Hit@1 & Hit@2 & Hit@3 & Hit@5 & Hit@10 & Hit@1 & Hit@2 & Hit@3 & Hit@5 & Hit@10 \\
    \midrule
    \single{} & 30.00 & 30.00 & 30.00 & 30.00 & 30.00 & 23.33 & 25.33 & 26.00 & 26.00 & 26.67 \\
    \dice{}-128 & 33.33 & 40.67 & 48.67 & 58.67 & 73.33 & 56.00 & 71.33 & 78.67 & 87.33 & 92.67 \\
    \dice{}-256 & 34.00 & 52.67 & 62.00 & 78.67 & 89.33 & 74.67 & 88.67 & 92.67 & 95.33 & 100.00 \\
    \dice{}-512 & 66.67 & 80.67 & 90.00 & 96.00 & 100.00 & 74.00 & 86.00 & 90.67 & 96.67 & 98.67 \\
    \dice{}-1024 & 90.00 & 100.00 & 100.00 & 100.00 & 100.00 & 74.00 & 88.00 & 90.67 & 94.00 & 98.67 \\
    \bottomrule
  \end{tabular*}
\end{table*}

\begin{table*}[h]
  \centering
  \caption{Dream adjacent single-vector reference under the same interface: \latechunkdoc{}. Synthetic scores are Hit@1 (\%), and real-task scores are nDCG@10. This probe situates \dice{} relative to a neighboring design that contextualizes token evidence before compression.}
  \label{tab:latechunk-doc}
  \small
  \begin{tabular*}{\textwidth}{@{\extracolsep{\fill}}lccccccccc}
    \toprule
    Method & Avg. & Pk$\leq$4k & Pk$>$4k & Nd$\leq$4k & Nd$>$4k & NQA & QMS & SFD & WQA \\
    \midrule
    \singlecite{} & 63.41 & \textbf{100.00} & 30.00 & 89.20 & 23.33 & 43.63 & 43.93 & 97.77 & 79.43 \\
    \latechunkdoccite{} & 81.51 & 96.00 & \textbf{93.33} & 90.00 & 73.33 & \textbf{65.05} & 50.13 & \textbf{98.53} & 85.69 \\
    \diceours{} & \textbf{81.92} & 98.80 & 90.00 & \textbf{92.40} & \textbf{74.00} & 65.01 & \textbf{50.54} & \textbf{98.53} & \textbf{86.04} \\
    \bottomrule
  \end{tabular*}
\end{table*}

\subsection*{B.1 Mistral Chunk-Size Sweep}
\label{app:mistral-chunk}

Table~\ref{tab:mistral-chunk-size} mirrors the Dream-side chunk-size ablation with a second backbone. Its role is to show that the main chunk-granularity trend remains visible on Mistral rather than being an artifact of a single encoder family. The same ordering appears here as in Dream: chunk-128 is too small on average, performance rises steadily with larger chunks, and chunk-1024 gives the strongest overall result together with the clearest gains on the difficult $>$4k synthetic slices.

\subsection*{B.2 Synthetic Top-\textit{k} Retrieval}
\label{app:topk-synth}

\begin{figure*}[h]
    \centering
    \includegraphics[width=\textwidth]{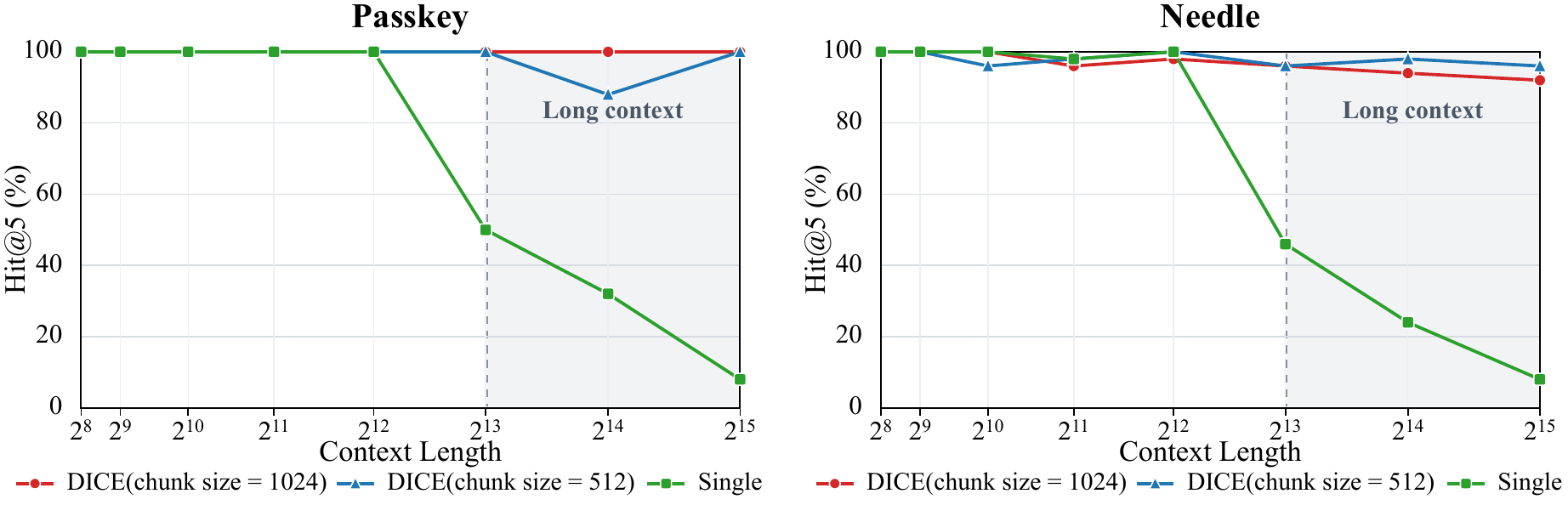}
    \caption{Per-length Hit@5 (\%) on synthetic tasks. The advantage of \dice{} remains concentrated in the longer contexts.}
    \label{fig:topk-length}
\end{figure*}

\begin{figure*}[t]
    \centering
    \includegraphics[width=\textwidth]{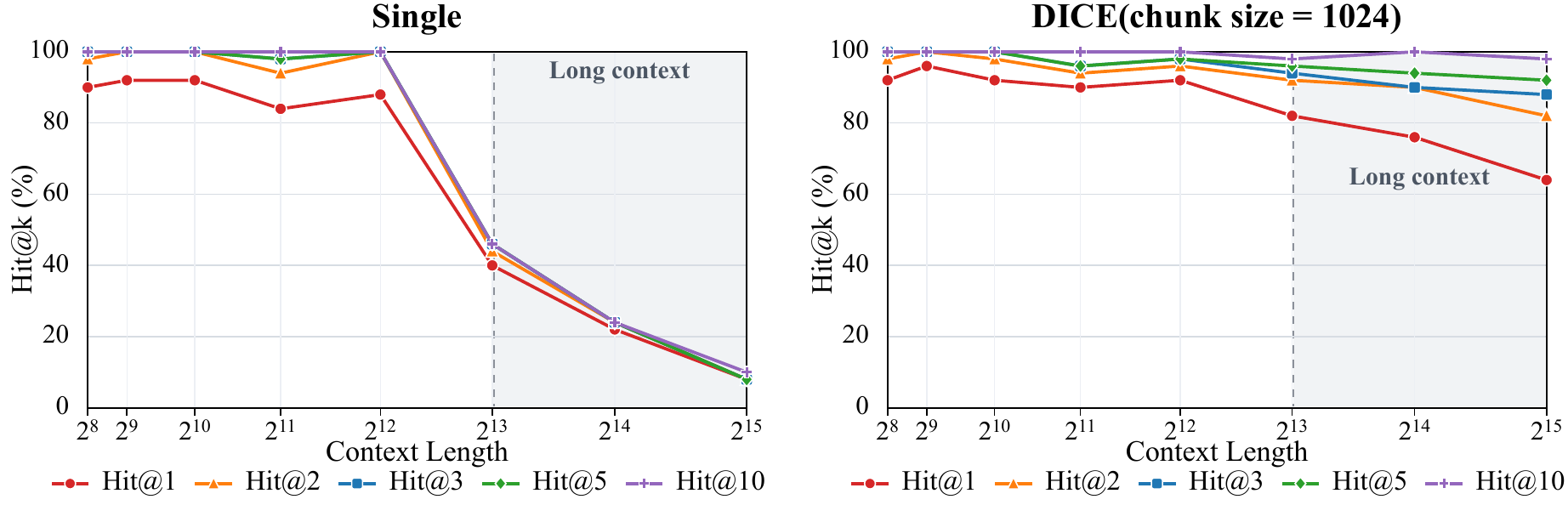}
    \caption{Needle Hit@$k$ (\%) by length. The full top-$k$ profile shows that \dice{} maintains its advantage as context length increases.}
    \label{fig:needle-topk-multiline}
\end{figure*}

\begin{figure*}[t]
    \centering
    \includegraphics[width=\textwidth]{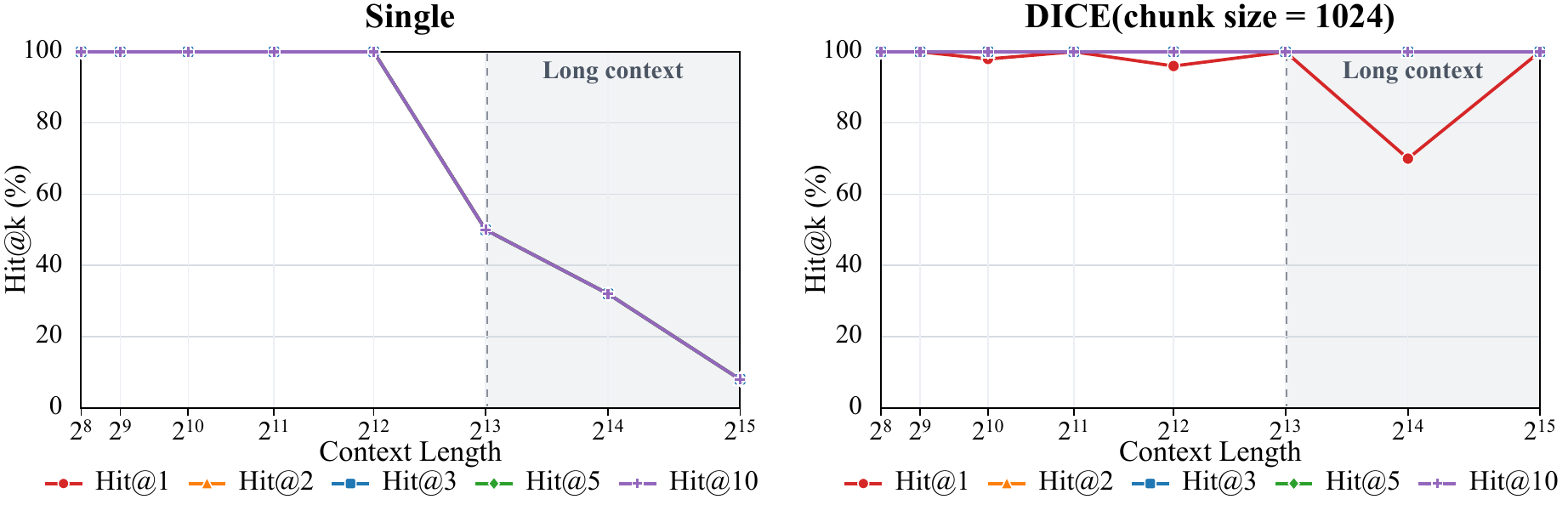}
    \caption{Passkey Hit@$k$ (\%) by length. Curve overlap here reflects substantive retrieval behavior rather than a plotting artifact.}
    \label{fig:passkey-topk-multiline}
\end{figure*}

These results complement Figure~\ref{fig:chunk-size} by showing that the same ranking trend persists beyond Hit@1 (\%). Table~\ref{tab:topk-synth} summarizes bucket-level Hit@$k$ (\%) on the hardest $>$4k synthetic slices, Figure~\ref{fig:topk-length} breaks Hit@5 (\%) down by length, and Figures~\ref{fig:needle-topk-multiline} and~\ref{fig:passkey-topk-multiline} show how the full Hit@$k$ (\%) profile evolves with length for Needle and Passkey, respectively. The improvement persists across broader top-$k$ cutoffs and remains most pronounced in the longest contexts, which is the pattern expected if chunk evidence is being preserved rather than accidentally re-ranked.

\begin{table*}[h]
  \centering
  \caption{Absolute-offset probe at chunk size 1024. Synthetic scores are Hit@1 (\%), and real-task scores are nDCG@10. Preserving absolute document offsets does not improve over the default local-reset position scheme.}
  \label{tab:abspos-probe}
  \small
  \begin{tabular*}{\textwidth}{@{\extracolsep{\fill}}llccccccccc}
    \toprule
    Backbone & Method & Avg. & Pk$\leq$4k & Pk$>$4k & Nd$\leq$4k & Nd$>$4k & NQA & QMS & SFD & WQA \\
    \midrule
    Dream & \dice{} & \textbf{81.92} & 98.80 & 90.00 & \textbf{92.40} & \textbf{74.00} & \textbf{65.01} & \textbf{50.54} & 98.53 & 86.04 \\
     & AbsPos & 81.85 & \textbf{99.20} & \textbf{90.67} & 90.80 & \textbf{74.00} & 64.98 & 50.28 & \textbf{98.54} & \textbf{86.34} \\
    \midrule
    Llama3 & \dice{} & \textbf{50.56} & \textbf{66.00} & \textbf{35.33} & \textbf{50.40} & \textbf{25.33} & \textbf{45.64} & 27.59 & \textbf{93.00} & \textbf{61.16} \\
     & AbsPos & 43.19 & 37.20 & 18.00 & 48.40 & 18.67 & 30.07 & \textbf{36.38} & 82.94 & 48.63 \\
    \midrule
    Mistral & \dice{} & \textbf{74.38} & 96.00 & \textbf{86.00} & \textbf{80.00} & \textbf{58.67} & \textbf{55.73} & \textbf{50.79} & \textbf{96.48} & \textbf{71.38} \\
     & AbsPos & 73.74 & \textbf{96.40} & 85.33 & 76.80 & 57.33 & 55.71 & 50.72 & 96.41 & 71.20 \\
    \bottomrule
\end{tabular*}
\end{table*}

\section*{Appendix C. Alternative Design References}

This appendix collects supplementary design probes that are adjacent to, but not part of, the main deployment-equivalent comparison. Their role is to sharpen the scope of the main claim: \dice{} is not presented as the only way to improve long-document retrieval, but as a particularly simple training-free design point within the one-query-one-document setting. The probes below clarify what happens when chunk evidence is contextualized differently or when local positional reset is removed.

\subsection*{C.1 Dream-Side Adjacent References}
\label{app:latechunk-doc}

Table~\ref{tab:latechunk-doc} reports one Dream-side adjacent reference under the same single-vector interface. \latechunkdoc{} contextualizes token evidence through a long-context forward pass before compressing it back into one document vector; it is included here as a neighboring design point rather than a deployment-equivalent primary baseline. The comparison is useful because it separates two ideas that are easy to conflate: delaying compression helps, but doing so through long-context contextualization is not the same design choice as independently encoding local chunks. The results suggest that richer contextualization before compression is indeed a strong neighboring design and can approach the performance of \dice{} while preserving a single stored vector per document.

\subsection*{C.2 Absolute-Offset Probe}
\label{app:abspos}

Table~\ref{tab:abspos-probe} compares the default local-reset position scheme with an absolute-offset variant that preserves each chunk's original document offset. This probe asks whether the gains of \dice{} come mainly from chunking itself or from the specific positional reset used in the default implementation. Across the completed backbones, absolute offsets never improve the average result and sometimes degrade it substantially, especially on Llama3. This makes the interpretation more specific: the benefit is not just that the document is split into pieces, but that each piece is encoded as a self-contained local context rather than as a late fragment in a very long positional range.

\end{document}